\documentclass[conference]{IEEEtran}
\IEEEoverridecommandlockouts
\renewcommand\thesection{\arabic{section}}
\renewcommand\thesubsection{\thesection.\arabic{subsection}}
\renewcommand\thesubsubsection{\thesubsection.\arabic{subsubsection}}
\renewcommand\thesectiondis{\arabic{section}}
\renewcommand\thesubsectiondis{\thesectiondis.\arabic{subsection}}
\renewcommand\thesubsubsectiondis{\thesubsectiondis.\arabic{subsubsection}}
\usepackage{cite}
\usepackage{amsmath,amssymb,amsfonts}
\usepackage{algorithmic}
\usepackage{graphicx}
\usepackage{textcomp}
\usepackage{xcolor}
\usepackage{enumitem}
\usepackage{float}
\usepackage{booktabs}
\usepackage{graphicx}
\usepackage{fancyhdr}
\usepackage{subcaption}
\usepackage{caption}
\usepackage[multiple]{footmisc}
\usepackage{hyperref}
\hypersetup{
    colorlinks=true,
    linkcolor=blue,
    filecolor=magenta,      
    urlcolor=blue,
    pdftitle={Sharelatex Example},
    bookmarks=true,
    pdfpagemode=FullScreen,
}
\let\oldbibliography\thebibliography
\renewcommand{\thebibliography}[1]{%
  \oldbibliography{#1}%
  \setlength{\itemsep}{0pt}%
}

\ifCLASSINFOpdf
\else
\fi
\begin{document}
%
\title{Flow-based Video Segmentation for Human Head and Shoulders}

\author{\IEEEauthorblockN{Zijian Kuang}
\IEEEauthorblockA{Department of Computing Science\\
University of Alberta\\
Edmonton, Canada \\
Email: kuang@ualberta.ca}
\and
\IEEEauthorblockN{Xinran Tie}
\IEEEauthorblockA{Department of Computing Science\\
University of Alberta\\
Edmonton, Canada \\
Email: xtie@ualberta.ca}
}


%


\maketitle
\thispagestyle{plain}
\pagestyle{plain}

\begin{abstract}
Video segmentation for the human head and shoulders is essential in creating elegant media for videoconferencing and virtual reality applications. The main challenge is to process high-quality background subtraction in a real-time manner and address the segmentation issues under motion blurs, e.g., shaking the head or waving hands during conference video. To overcome the motion blur problem in video segmentation, we propose a novel flow-based encoder-decoder network (FUNet) that combines both traditional Horn-Schunck optical-flow estimation technique and convolutional neural networks to perform robust real-time video segmentation. We also introduce a video and image segmentation dataset: ConferenceVideoSegmentationDataset. Code and pre-trained models are available on our GitHub repository: \url{https://github.com/kuangzijian/Flow-Based-Video-Matting}.
\end{abstract}


%
\IEEEpeerreviewmaketitle

\section{Introduction}
Video segmentation is widely used in movie editing, virtual reality applications, and video conferencing applications like Zoom, Google meet, and Microsoft Teams \cite{bgmv2}. Besides the entertainment purpose, such as the post-production of movies or virtual reality scenes, the video segmentation can also protect people’s privacy during video conferencing by hiding the location and environment behind the users \cite{bgmv2}.

\begin{figure}[H]
\centerline{\includegraphics[width=3.5in]{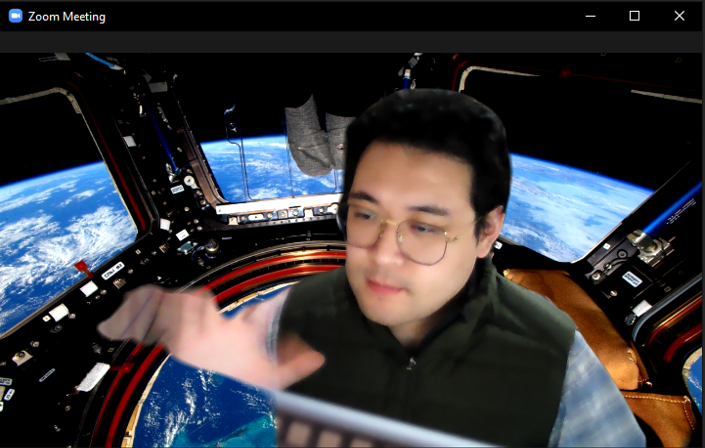}}
\caption{Zoom's background replacement application cannot detect waving hand accurately}
\label{zoom}
\end{figure}

The traditional Gaussian Mixture model-based background subtraction algorithms such as GMG, KNN, MOG and MOG2 \cite{gmg} \cite{knn} \cite{mog} \cite{mog2} assume a moving object in the foreground to a static background \cite{tbb}. However, if the user stops moving in the video, then the traditional background subtraction methods will fail. On the other hand, most deep learning-based video segmentation requires a huge amount of training data and some models even require a clean background image as input before performing prediction \cite{bgmv2} \cite{jbs}. The video segmentation with motion blur is also challenging that even the state-of-the-art learning-based model cannot handle appropriately \cite{MODNet}; for example, video segmentation is not accurate when people are shaking the head or waving their hands in most of the conferencing applications as show in Fig. \ref{zoom}.

In this paper, a novel flow-based encoder-decoder network(FUNet) is proposed to detect a human head and shoulders from a video and remove the background to create elegant media for videoconferencing and virtual reality applications. The proposed model combines both traditional Horn-Schunck optical-flow-based technique \cite{pwc} and convolutional neural networks to perform robust real-time video segmentation. In our model, an optical-flow-based model is utilized to extract motion features between every two frames. Then we combine both the motion feature and the appearance feature from the original frame and utilize an encoder-decoder network to learn and predict a mask output for human head and shoulders segmentation from the background. The convolutional neural networks are implemented to speed up the video segmentation process to enable processing video frames in a real-time manner.

\section{Related Work}
\label{RW}
Background replacement aims to extract the foreground from the input video and then combine it with a new background. The methodologies of background replacement can be divided into matting, or segmentation \cite{bgmv2}. Image or video matting can perform visually detailed composites but requires very detailed manual annotations. While the background segmentation focuses on labeling the pixels between the foreground objects and background, its performance is much faster and more efficient \cite{bgmv2}. In this section, we discuss related works regarding background replacement methods and related methods for potential improvements on segmentation methods.

\subsection{Image and Video Matting}
Image and video matting are estimating the foreground objects accurately. Unlike image or video segmentation generates a binary image by labeling the foreground and background pixels, the matting method can handle those pixels that may belong to the foreground and background, called the mixed pixels.

Background Matting V2 (BGM V2) is one state of the art networks introduced to overcome the challenge that users do not typically have access to green screens to replace the background during a video conferencing. Consequently, there is residual of the original background shown in the close-ups of users’ hairs and glasses. In the paper \cite{bgmv2}, S. Lin et al. proposed to employ two neural networks to achieve real-time, fully automated, and high-resolution background matting. One neural network in the architecture is worked as the base network by providing an extra background image. The other neural network is served as the refinement network. The base network is responsible for predicting the alpha matte and foreground layer at a lower resolution. After that, an error prediction map specifies the areas that need refinement to be passed to the refinement network. Then, the refinement network takes the low-resolution result and the original image to refine the regions with significant predicted errors. Lastly, the alpha and the foreground residuals are produced at the original resolution \cite{bgmv2}.

Another state-of-the-art image and video matting algorithm is called MODNet \cite{MODNet} introduced by Z. Ke et al. in 2020. The proposed MODNet takes only one RGB image as the input and then decomposes the matting process into three correlated sub-tasks to learn and predict simultaneous. The three sub-tasks include predicting human semantics $s_{p}$, boundary details $d_{p}$, and final alpha matte $\alpha_{p}$ through three separate pipelines S, D, and F \cite{MODNet}.  To predict the coarse semantic sp, the model uses L2 loss under supervised learning using low-resolution representation S(I) against a thumbnail of the ground truth matte $\alpha_{g}$. The equation is formed as below:

\begin{equation}
\mathcal{L}_{s}=\frac{1}{2}\left\|s_{p}-G\left(\alpha_{g}\right)\right\|_{2}
\end{equation}

The detailed boundary feature is calculated using D(I, S(I)) and L1 loss function as below:

\begin{equation}
\mathcal{L}_{d}=m_{d}\left\|d_{p}-\alpha_{g}\right\|_{1}
\end{equation}
Where $m_{d}$ is a generated binary mask through dilation and erosion to let ${L}_{d}$ focus on the boundaries \cite{MODNet}.

Finally, the model combines both semantics and details by concatenating S(I) and D(I, S(I)) to predict the final alpha matte $\alpha_{p}$ using:

\begin{equation}
\mathcal{L}_{\alpha}=\left\|\alpha_{p}-\alpha_{g}\right\|_{1}+\mathcal{L}_{c}
\end{equation}

\subsection{Image and Video Segmentation}
Image segmentation aims to cluster pixels into salient image regions such as regions corresponding to individual surfaces, objects, or natural parts of objects.

U-Net \cite{unet} is a fully convolutional network proposed for the problem of biomedical segmentation problem. The architecture of this network is modified to work with few training samples and produce more precise segmentation. A usual contracting layer is supplemented with successive layers, and the upsampling operators are used instead in the network. Features from the contracting path are combined with upsampled output to boost the accuracy of the localization better. In order to assemble a more precise output, a successive convolution layer is used for learning.
\begin{figure*}[!hbt]
    \centerline{\includegraphics[width=7in]{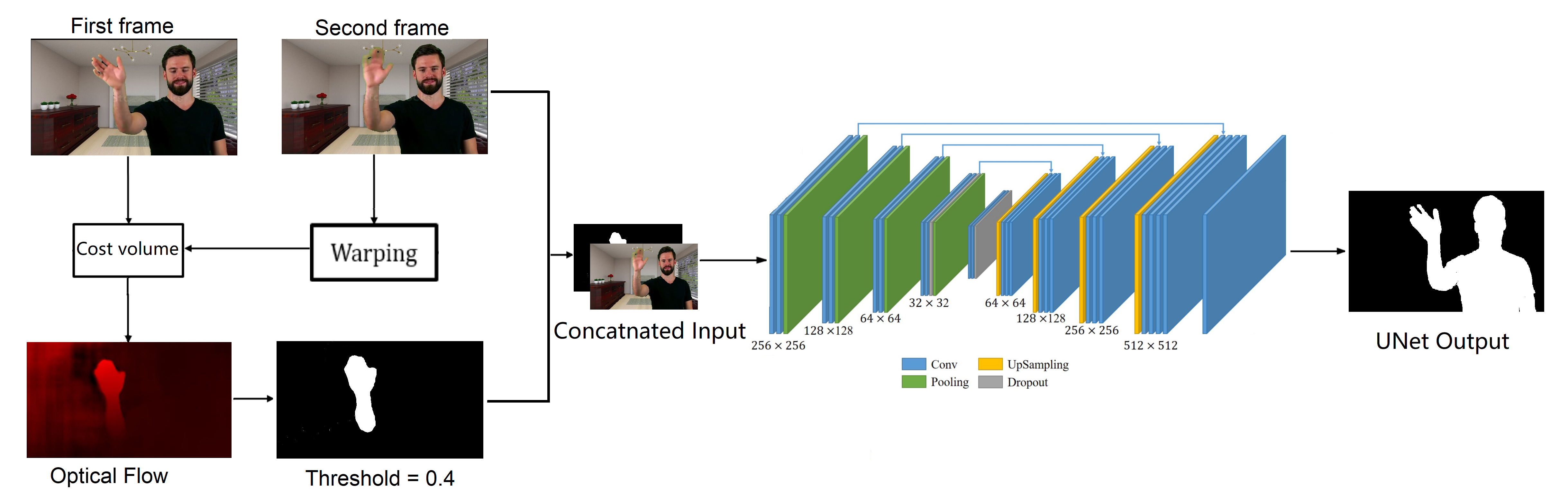}}{}
    \caption{Overview of our proposed network. It first estimates optical flows for every two frames, and extract the magnitude information as motion feature $\left[M_{1}, M_{2}, \ldots, M_{N}\right]$ for each frame (Section 3.1). Then we fuse the motion feature and the appearance feature from the original frame and feed it into an encoder-decoder architecture to perform video segmentation (Section 3.2)}
    \label{network}
\end{figure*}
Most importantly, many feature channels are used in the upsampling part for propagating context information to high-resolution layers. Moreover, excessive data argumentation is applied due to a small amount of available training data. During training, both input images and the corresponding segmentation images are used as input. The soft-max is defined as 

\begin{equation}
p_{k}(\mathbf{x})=\exp \left(a_{k}(\mathbf{x})\right) /\left(\sum_{k^{\prime}=1}^{K} \exp \left(a_{k^{\prime}}(\mathbf{x})\right)\right),
\end{equation} 

where $a_{k}(\mathbf{x})$ represents the activation in feature channel $k$ at the pixel position $x$. To penalize the deviation of $p_{\ell(\mathbf{x})}(\mathbf{x})$ from 1 for each position, the cross-entropy is defined as 

\begin{equation}
E=\sum_{\mathbf{x} \in \Omega} w(\mathbf{x}) \log \left(p_{\ell(\mathbf{x})}(\mathbf{x})\right),
\end{equation}

Where $\ell$ is the ground truth for each pixel and $w$ is a weight map. After computing the separation border, the weight map is computed as 

\begin{equation}
w(\mathbf{x})=w_{c}(\mathbf{x})+w_{0} \cdot \exp \left(-\frac{\left(d_{1}(\mathbf{x})+d_{2}(\mathbf{x})\right)^{2}}{2 \sigma^{2}}\right),
\end{equation}

Where $w_{c}$ is the weight map, $d_{1}$ is the distance to the border of the nearest cell, and $d_{2}$ is the distance to the border of the second nearest cell.

A. Basu, I. Cheng et al. introduced a methodology for airways segmentation using Cone Beam CT data and measuring airways' volume. Since many of the researches focus on the segmentation and reconstruction of lower airway trees, a new strategy for detecting the boundary of slices of the upper airway and tracking the airway's contour using Gradient Vector Flow (GVF) snakes is introduced. 

The traditional GVF snakes are not performing well for airway segmentation when applied directly to CT images. Therefore, this paper's new method has modified the GVF algorithm with edge detection and sneak-shifting steps. By applying edge detection before the GVF snakes and using snake shifting techniques, the prior knowledge of airway CT slices is utilized, and the model works more robustly. The previous knowledge of the shape of the airway can automatically detect the airway in the first slice. The detected contour will then be used as the second slice's sneak initialization and so on \cite{airway}. A heuristic is also applied to differentiate bones from the airway by the color to make sure the snake converges correctly. Following this, the airway volume is estimated based on the 3D model constructed with automatically detected contours.

\subsection{Optical Flow}
Optical flow is a pattern of apparent motion of image objects between two consecutive frames caused by the movement. The pattern is represented in the 2D vector field. 

PWC-Net \cite{pwcnet} is a CNN model that is proposed to use the principles of pyramidal processing, warping, and cost volumes for optical flow. This model is 17 times smaller in size, two times faster in inference, much easier to train, and outperforms all the published optical flow models on two of the benchmarks. In the paper, PWC-Net firstly uses the optical flow estimate to warp the features extracted from the second image. Then, it utilizes both of the warped features and the features from the first image to construct a cost volume. At last, the cost volume is processed by the network to predict the optical flow. In the first stage, a learnable feature L-level pyramid is constructed with raw images as the input. After that, PWC-Net warps the features of the second image towards the first image using the upsampled flow from the $l+1$th level:

\begin{equation}
\mathbf{c}_{w}^{l}(\mathbf{x})=\mathbf{c}_{2}^{l}\left(\mathbf{x}+\mathrm{up}_{2}\left(\mathbf{w}^{l+1}\right)(\mathbf{x})\right).
\end{equation}

Then, the warped features along with the features of the first image are used to compute the cost volume. A cost volume is utilized to store the matching cost of associating two pixels in two frames. The matching cost between the two sets of features is defined as 

\begin{equation}
\mathbf{c v}^{l}\left(\mathbf{x}_{1}, \mathbf{x}_{2}\right)=\frac{1}{N}\left(\mathbf{c}_{1}^{l}\left(\mathbf{x}_{1}\right)\right)^{\top} \mathbf{c}_{w}^{l}\left(\mathbf{x}_{2}\right),
\end{equation}

where $\top$ is a transpose operator and N is the length of the column vector $\mathbf{c}_{1}^{l}\left(\mathbf{x}_{1}\right)$. Next, the cost volume and original image features, and upsampled features are processed by the multi-layer optical flow estimator. At last, a context network is used to post-process the flow.

\section{Proposed Method}
In this section, we elaborate on the architecture of FUNet and discuss more details of the network design, as shown in Fig.\ref{network}. Our goal is to detect a human head and shoulders from a video and remove the background to create elegant media for videoconferencing and virtual reality applications. We proposed a network to predict a binary mask for video segmentation: given a sequence of video frames $\left[I_{1}, I_{2}, \ldots, I_{N}\right]$, our model can label the pixel as foreground or background in each frame, and predict a sequence of binary masks $\left[O_{1}, O_{2}, \ldots, O_{N}\right]$.

We first estimate optical flows for every two frames, and extract the magnitude information as motion feature $\left[M_{1}, M_{2}, \ldots, M_{N}\right]$ for each frame (Section 3.1). Then we fuse the motion feature and the appearance feature from the original frame and feed it into an encoder-decoder architecture that inspired from UNet \cite{unet} to perform video segmentation (Section 3.2).

\subsection{Motion Feature Extraction}
Since the optical flow is a pattern of apparent motion of image objects between two consecutive frames caused by the movement, inspired by the PWCNet \cite{pwc}, we use the same learnable feature pyramids and layer-wise warping operation along with cost volume calculation to estimate optical flow. 

For two images input $I_{1}$ and $I_{2}$, the model generates L-level pyramids of extracted features from each frame. The model warps $I_{2}$'s features toward $I_{1}$'s features at each lth level. Then the model constructs a cost volume to store the matching costs between each pixel in the current frame with its corresponding pixels in the previous frame.

With given sequence of video frames $\left[I_{1}, I_{2}, \ldots, I_{N}\right]$, the motion feature extraction process can output a sequence of optical flow vectors $\left[v_{1}, v_{2}, \ldots, v_{N}\right]$ for each frame, we further formed a indicator function $\mathbb{I}(x)$ to separate the optical flow vector value into either foreground or background using a threshold $\alpha$:

\begin{equation}
\mathcal M_{n}(x)=\left\{\begin{array}{ll}
1 & \text { for } v_{n}(x) \geq \alpha \\
0 & \text { for } v_{n}(x)<\alpha
\end{array}\right.
\end{equation}

Where $v_{n}\left(\mathbf{x}\right)$ denotes the optical flow magnitude for pixel x in $n$th frame. The $M_{n}$ is the extracted motion feature mask for $n$th frame.

\subsection{Motion and Appearance Fusion}
After retrieving the sequence of extracted motion feature masks $\left[M_{1}, M_{2}, \ldots, M_{N}\right]$ from the motion feature extraction process (Section 3.1), we further concatenate the motion feature M and original image's appearance feature I, and pass the concatenated value into an encoder-decoder structure to perform video segmentation. 

The architecture consists of an encoder and a decoder part. The encoder part is built with the repeated application of two 3x3 convolution layers, each followed by a ReLU and a 2x2 max pooling operation with downsampling using stride 2. The decoder part consists of an upsampling along with a 2x2 convolutional layer. The model also includes a concatenation (shown as light blue lines on top of the model in Fig. \ref{network}) from the corresponding cropped feature map from the encoder side and two 3x3 convolutions, each followed by a ReLU.

We use binary cross-entropy loss with a pixel-wise softmax (BCE With logits Loss) as our loss function:

\begin{equation}
L = -\frac{1}{n} \sum\left(O_{n} \times \ln C_{n}+\left(1-O_{n}\right) \times \ln \left(\mathbf{1}-C_{n}\right)\right)
\end{equation}

Where $C_{n}$ denotes the predicted probability of the $n$th pixel, and $O_{n}$ denotes the ground-truth value of the $n$th pixel.

\section{Implementation Details}

\subsection{Training}
Our model is trained on eight video sequences that contain 2600 frames. During training, optical flows between every two frames are firstly estimated. Since there is no previous frame for the first frame, the first optical flow generated using the first and second frames is duplicated to keep the number of optical flow images consistent with the number of training images. Next, the estimated optical flow magnitude is computed and is later split into the foreground and background under a threshold, which resulted in a binary mask as our motion feature layer. We found the threshold $\alpha$ = 0.4 works best to separate foreground and background during our experiments. Then, we concatenate the motion feature mask as the fourth channel into the original image and feed it to the convolutional neural network. 

The model is implemented with the BCE With Logits Loss and the Root Mean Square Propagation (RMSProp) optimizer. RMSProp is a gradient-based optimization technique that uses an adaptive learning rate and a moving average of squared gradients to normalize the gradient. During training, the learning rate is set to 0.0001 with a weight decay of 1e-8, and the momentum equals 0.9. BCE With Logits Loss combines a plain sigmoid after the Binary Cross Entropy between the target and the output in a single class, promoting numerical stability for training. 

\subsection{Dataset}
\begin{figure}[!hbt]
\centerline{\includegraphics[width=3.5in]{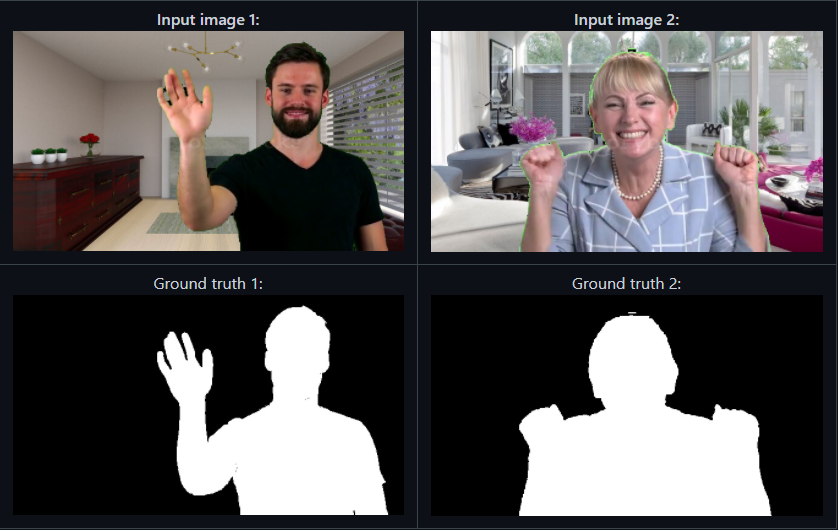}}
\caption{Examples of video frames and ground truth masks of our ConferenceVideoSegmentationDataset}
\label{dataset}
\end{figure}

We created our own video segmentation dataset. The source data includes ten online conference-style green screen videos. We extracted 3600 frames from the videos and generated the ground truth masks for each character in the video, and then we applied virtual background to the frames as our training/testing dataset. The examples of our dataset are shown in Fig.\ref{dataset}. The dataset is available for download under our GitHub repository\footnote{https://github.com/kuangzijian/Flow-Based-Video-Matting}.

\section{Experiments Results}
Input images and their corresponding segmentation masks are used as input during training. The model is implemented based on the PyTorch framework and is trained on a piece of RTX 2080 Ti GPU. A total of 10 epochs is trained for the model with a learning rate of 0.0001 and a batch size of 1. Original images of the dataset are used for training with an available down scaling factor that could be adjusted in future experiments and studies. BCE With Logits Loss is used as the loss function for the model to minimize the error between the predicted mask and its ground truth. 

\subsection{Evaluation Method}
During validation and testing, the Dice coefficient (F1 score) is used to evaluate the performance of the proposed network. Dice coefficient calculates the size of the overlap of the two segmentation masks divided by the total size of two masks resulted in a value range from 0 to 1. A higher Dice coefficient indicates a better performance which detects a greater similarity between the prediction and the ground truth. In our experiments, 60\% of the dataset is used for training, and 20\% is used for validation. The rest 20\% of the dataset is used for testing.

\subsection{Experiment – Evaluate the video segmentation on our testing dataset}
This experiment evaluated the proposed model on our testing dataset, which contains two video sequences consisting of 720 frames. In each video sequence of our testing dataset, the background image and foreground human are unique and different from our training dataset. The overall performance on the segmentation is very promising as the averaging Dice coefficient (F1 score) on these testing video sequences is 0.96. The example frames of our evaluation results are shown as follow:

\begin{figure}[H]
\centerline{\includegraphics[width=4.5in]{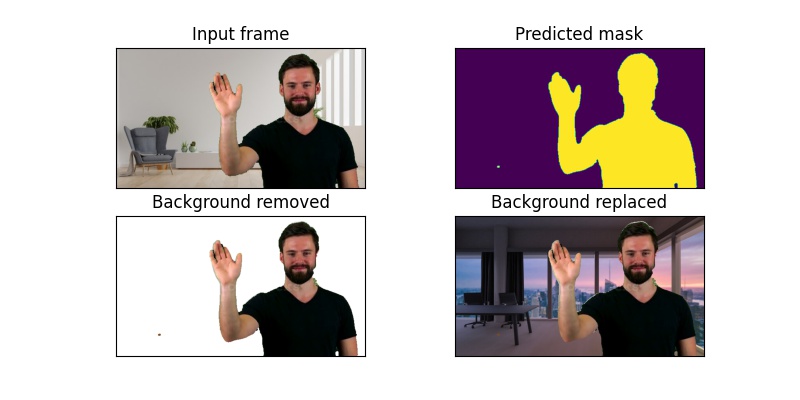} 
}
\centerline{\includegraphics[width=4.5in]{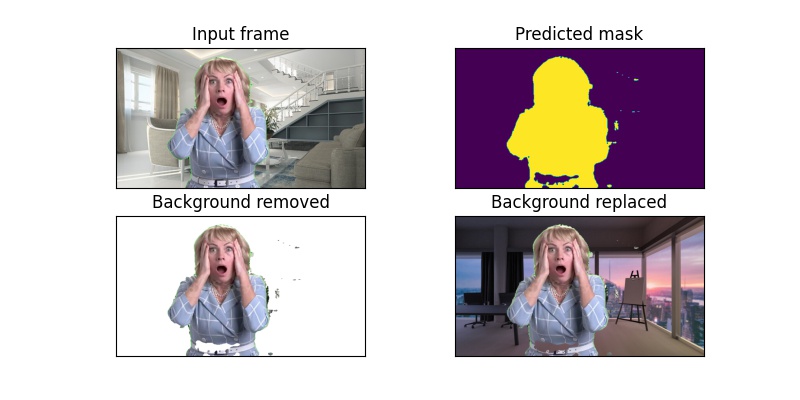} 
}
\caption{Evaluation on our testing dataset with an average Dice coefficient (F1 score) of 0.96 as the result}
\label{eva1}
\end{figure}

\subsection{Future Experiment – Evaluate the video sequence on real-world data}
In the future experiment, we plan to evaluate the proposed model on the real-world data recorded from conference video applications, such as Zoom, Google Meeting, and Microsoft Teams. Due to time constraints, we are still working on data collection and data preprocessing for evaluation. The real-world dataset and detailed experiment results will be coming soon. Also, due to time constraints, we have not gotten a chance to compare with other models on a benchmark dataset, which we will do in the future experiments.

\section{Conclusion and Future Work}
In conclusion, we proposed a novel flow-based encoder-decoder network to detect a human head and shoulders from a video and remove the background to create elegant media for videoconferencing and virtual reality applications. We also created our own conference video style segmentation dataset called ConferenceVideoSegmentationDataset for further studies and researches.

In the future, we would like to improve our model by implementing a fully convolutional encoder-decoder network inspired by the DeepLabV3 \cite{deeplabv3}. We will also improve our predicting speed with the softmax splatting method to warp and generate the masks in between every ten frames so that our model will only need to predict every first and tenth frame interpolation technique to predict the missing masks. Most importantly, we will collect more conference-style data to expand our dataset to improve our model's performance.

\section*{Acknowledgment}
The authors would like to thank our mentor Xuanyi Wu, for her guidance and feedback throughout the research and study. We would also thank our advisor Dr. Anup Basu for his motivation and support to bring out the novelty in our research.

\bibliographystyle{IEEEtran}
\bibliography{main}
\vspace{-1 cm}
\begin{IEEEbiographynophoto}{Zijian Kuang}
is a machine learning enthusiast who is currently pursuing his MSc. in Computing Science at the University of Alberta. He has 5+ years of work experience in project management and software development for government and large organizations.
\end{IEEEbiographynophoto}

\vskip 0pt plus -1fil
\begin{IEEEbiographynophoto}{Xinran Tie}
is currently a student pursuing MSc. in Computing Science with Specialization in Multimedia at the University of Alberta. She is expected to graduate in December 2021, and currently studies on and researches 3 projects from the courses. 
\end{IEEEbiographynophoto}

\end{document}